\documentclass[10pt,journal,compsoc]{IEEEtran}

\usepackage{amsfonts, bbm}
\usepackage{amsmath}
\usepackage{enumerate}
\usepackage{enumitem}
\usepackage{graphicx}
\usepackage{hyperref}
\usepackage[all]{xy}
\usepackage{wrapfig}
\usepackage{fancyvrb}
\usepackage{listings}
\usepackage{amsthm}
\usepackage{centernot}
\usepackage{mathtools}
\usepackage{verbatim}
\usepackage{subcaption}
\usepackage[numbers]{natbib}
\DeclareMathOperator*{\argmax}{arg\,max}

\ifCLASSOPTIONcompsoc
  \usepackage[nocompress]{cite}
\else
  \usepackage{cite}
\fi
\ifCLASSINFOpdf
\else
\fi
\begin{document}
\title{Vertex Classification on Weighted Networks}

\author{Hayden Helm, %~\IEEEmembership{Member,~IEEE,},
        Joshua T. Vogelstein
        and Carey E. Priebe %, \IEEEmembership{Senior Member, IEEE}
\thanks{H.H. is with the Center for Imaging Sciences at Johns Hopkins University; J.T.V. is with the Department of Biomedical Engineering at Johns Hopkins University; C.E.P. is with Department of Applied Mathematics
and Statistics at Johns Hopkins University. }}% <-this % stops a space

\IEEEtitleabstractindextext{%
\begin{abstract}
This paper proposes a discrimination technique for vertices in a weighted network. We assume that the edge weights and adjacencies in the network are conditionally independent and that both sources of information encode class membership information. In particular, we introduce a edge weight distribution matrix to the standard K-Block Stochastic Block Model to model weighted networks. This allows us to develop simple yet powerful extensions of classification techniques using the spectral embedding of the unweighted adjacency matrix. We consider two
% TODO: assumptions are only relevant for theory, so, no need to mention them outside the context of a theoretical result
assumptions on the edge weight distributions and propose classification procedures in both settings. We show the effectiveness of the proposed classifiers by comparing them to quadratic discriminant analysis following the spectral embedding of a transformed weighted network. Moreover, we discuss and show how the methods perform when the edge weights do not encode class membership information. 
\end{abstract}

}

\maketitle

\IEEEdisplaynontitleabstractindextext

% \fi
\IEEEpeerreviewmaketitle

\IEEEraisesectionheading{\section{Introduction}\label{sec:introduction}}

% TODO: add citations
\IEEEPARstart{W}{eighted} networks are common in many research fields ranging from neuroscience to sociology. While networks provide a rich source of information, it can be difficult to identify patterns and groupings within the data. 
Hence, problems that require understanding relationships within and across groups of nodes, which we will refer to as communities or classes, are non-trivial.
For vertex, or node, classification the objective is to predict the class label for each node where we assume that a node belongs to exactly one of K classes. In a neuroscience application, for example, the classes may represent types of neurons. %\citep{hall1991posterior}.
% TODO: citation

One way to address the vertex classification problem is by finding a low-dimension Euclidianal representation of the unweighted network and subsequently
% TODO: do you mean "and composing with" rather than "using"?
using common discrimination techniques, such as k-nearest neighbors, on the transformed data \citep{sussman2014consistent}.
% TODO: think of citation numbers as footbotes, they cannot be part of a sentence.
A similar results shows universal consistency for this type of procedure for a very general class of unweighted network models \citep{tang2013universally}. 

Blindly applying these methods to weighted networks, however, is ineffective in simulation and practice. This is likely due to noise introduced by edge weights.
% TODO: "noise" or "signal ignored"?
Normalizing the weights before embedding the network, as pass-to-ranks does, mitigates the effect of this noise on subsequent inference. 
% TODO: explaining PTR in detail here is not necessary. put technical stuff, anything with notation, outside the intro.  
As the name suggests, pass-to-ranks uses the rankings of the edge weights to transform a weighted adjacency matrix from $ A \in \mathbb{R}^{n \times n} $ to $ A_{ptr} \in ([0,2] \cap \mathbb{Q})^{n \times n} $ by changing the value of the edge weight. The new weight is equal to two times the rank of the original edge of $ A $ divided by $ |E| $, the size of the edge set. Though pass-to-ranks is useful for node classification, it is hard to pin down analytically due to the method's minimal assumptions on the edge weights. One of the goals of the this paper is to introduce a more tractable framework for effective node classification on weighted networks. 

% TODO: outline paragraphs i find just add text.  i can look at the section headers if i want. think about anything you've ever read by a professional writer.  they never have a paragraph outline.

% Following this introduction, Section 2 provides necessary preliminary information. Sections 3 and 4 present new methods for node classification in a weighted network, showcase the effectiveness of the proposed methods, and discuss their sensitivity to misspecification. In Section 3 we classify in a setting where it is assumed that the weight distributions are ordered. In section 4 we classify nodes in a more general setting. Afterwards, Section 5 applies the proposed method on a C. elegans connectome. Section 6 discusses method limitations and areas for further study.

\subsection{Problem Statement}
It is important to completely characterize the problem we address before we continue. We use notation and concepts here that are explained in more detail in later sections. 

Our goal is to classify unlabeled nodes in a weighted network. In general, we are given a weighted network, $ G = (V, E) $, where $ V $ is a set of nodes and $ E $ is a set of edges. Note that $ (i, j, w_{ij}) \in E $ if the edge between node $ i $ and node $ j $ exists and has weight $ w_{i,j} $. In our setting we deal with symmetric (if $ (i, j, w_{i,j}) \in E $ then $ (j, i, w_{i,j}) \in E) $ and hollow $ (i, i, w_{i,i}) \notin E $) networks. 
% TODO: avoid paragraphical stuff like the above, make the sentence flow.

We represent this network as a weighted adjacency matrix, denoted $ C $, where we think of $ C $ as the Hadamard, or entrywise, product of the unweighted adjacency matrix $ A $ and the matrix of weights $ W $. That is, $ C = A \circ W $. Additionally, we are given a set of nodes with known class membership, referred to as training or labeled nodes, that we use to inform our procedure. In this paper there's an explicit assumption that $ W $ encodes block membership information. 
% TODO: it is not at all clear what it means to say that W encodes something.  in particular, encoding is a statistical concept, and there are no random variables defined yet, etc. 
There exists powerful methods for dealing with $ A $, outlined in section 2.3, and so our focus will be on handling $ W $ and, in turn, $ C $.

Hence, this paper is an exploration of how we can use the class membership information encoded in $ W $ to more accurately classify unlabeled nodes. Specifically, we assume that there is a symmetric $ K \times K $ matrix of distributions, $ \mathcal{F} $, where the $ (u, v)^{th} $ entry of $ \mathcal{F} $ is the distribution governing the edge weights between the nodes in block $ u $ and the nodes in block $ v $ (Sections 2.2 and 2.4.1). We estimate these distributions using the edge weights between the training nodes in block $ u $ and the training nodes in block $ v $. The estimated distributions are denoted $ \hat{\mathcal{F}}_{u, v} $. Consequently, for block $ u $, we have a vector of estimated distributions $ \hat{\mathcal{F}}_{u} = (\hat{F}_{u, 1}, \hdots, \hat{F}_{u, K}) $. 

Note that we observe the edge weights between an unlabeled node and the training nodes for each block. Hence, for a particular unlabeled node $ i $ we can estimate the distributions corresponding to each block. That is, $ \hat{\mathcal{F}}(i) = (\hat{\mathcal{F}}(i)_{1}, \hdots, \hat{\mathcal{F}}(i)_{K}) $ for unlabeled node $ i $. Extracting class membership information for the unlabeled node from this collection of vectors comes down to comparing $ \hat{\mathcal{F}}(i) $ to each of the $ \hat{\mathcal{F}}_{u} $. 

Letting $ \hat{\mathcal{F}}_{u,v} $ be the empirical cumulative distribution is perhaps the most general treatment of the edge weight distributions and is addressed in Section 4. We explore a more restrictive model in Section 3.

\section{Preliminaries}

\subsection{Stochastic Block Model}
The network model used in this paper is the Stochastic Block Model (SBM), which is a restricted version of the Random Dot Product Graph (RDPG) \citep{young2007random}. An RDPG is an independent edge random graph that is characterized by a collection of positions in $ \mathbb{R}^{d} $ that correspond to the nodes in the network. In particular, each node $ i $ in the network has a "position", $ X_{i} \in \mathbb{R}^{d} $ where the only restriction on $ X_{i} $ is that $ \langle X_{i}, X_{j} \rangle \in [0, 1] \; $ for all $ i,j $, where $ \langle \cdot, \cdot \rangle $ is the dot product of two vectors. The SBM is an RDPG where $ X_{i} \in \{X_{1}, \hdots, X_{K}\} $, where $ K $ is the number of blocks or classes. 

In an SBM the probability that an edge exists between two nodes depends only on the class memberships of the nodes. Importantly, the true positions are typically unknown and are referred to as latent positions. We call the estimates of the latent positions estimated positions.

The SBM is a common generative model used for network analysis because of its simple description and ability to capture complex network structures (see \citep{abbe2017community} sections 1 and 2 for history and literature overview and \citep{bickel2013asymptotic} for analysis of parameter estimation techniques). Four objects completely describe the model. The number of blocks in the network, $ K $. The set of nodes, V, where $ |V| = n $. The (sometimes partially observed) block membership function $ b: V \rightarrow [K] $ which implies block membership priors $ \pi = (\pi_{1}, \pi_{2}, .., \pi_{K}) \in \Delta_{K - 1} $. And, finally, the matrix that governs adjacency information
$$ B = \begin{bmatrix}
\langle X_{1}, X_{1} \rangle & \hdots & \langle X_{1}, X_{K} \rangle \\
\vdots & \ddots & \vdots \\
\langle X_{K}, X_{1} \rangle & \hdots & \langle X_{K}, X_{K} \rangle
\end{bmatrix} $$ where $ X_{u} $ is the latent position corresponding to nodes in block $ u $. The existence of an edge between node $ i $ and node $ j $, where $ b(i) = u $ and $ b(j) = v $, is generated from a coin flip with weight equal to $ B[b(i), b(j)] = B_{u,v} $.

The analysis in this paper is focused on 2 block matrices of the form 
$$ B = \begin{bmatrix} p^{2} & pq \\ pq & q^{2} \end{bmatrix} $$ Notice that $ det(B) = p^{2}q^{2} - p^{2}q^{2} = 0 $. Using the characteristic equation to find the eigenvalues, 
\begin{align*}
    \lambda^{2} - (p^{2} + q^{2})\lambda  &= 0 \\
    \lambda(\lambda - p^{2} - q^{2}) &= 0 \\
    \lambda_{1}, \lambda_{2} &= 0, p^{2} + q^{2}
\end{align*} So $ rank(B) = 1 $ if $ p > 0 $ or $ q > 0 $. We assume that $ rank(B) = 1 $ is known throughout this paper. Otherwise, estimating $ rank(B) $ is a complicated task in and of itself \citep{zhu2006automatic}.
% TODO: estimating rank(B) is complicated regardless of assumptions.  so, not clear what you mean here.

\subsection{Adjacency Spectral Embedding}
The method underlying the results of consistent vertex classification (as in \citep{sussman2014consistent}, \citep{tang2013universally}) is the Adjacency Spectral Embedding (ASE) of a network. ASE transforms the network into a collection of objects in Euclidian space using the Singular Value Decomposition (SVD). In particular, $$ A = U \Sigma U^{T} $$ where $ U $ is orthogonal and $ \Sigma $ is a diagonal matrix with the the singular values of $ A $ occupying the diagonals in decreasing order. For  notational simplicity we will assume $ B $ is positive semi defininte.
% TODO: A or P?
\citep{athreya2016limit} shows that if $ A $ is generated from an RDPG then the rows of $ U\Sigma^{1/2} $ are asympotically normally distributed around an orthogonal transformation of the latent positions that generated $ B $. See \citep{von2007tutorial} for an implementation tutorial and \citep{athreya2017statistical} for a survey of results on spectral embeddings of RDPGs.

\begin{comment}
For the two block rank one case
\begin{align*}
    X_{i} &\sim \mathcal{N}\bigg(p, \frac{\pi_{1}p^4(1 - p^2) + \pi_{2}pq^3(1 - pq)}{n(\pi_{1}p^2 + \pi_{2}q^2)^2}\bigg) \mbox{ if }b(i) = 1 \\
    \\
    X_{i} &\sim \mathcal{N}\bigg(q, \frac{\pi_{1}p^3q(1 - pq) + \pi_{2}q^4(1 - q^2)}{n(\pi_{1}p^2 + \pi_{2}q^2)^2}\bigg) \mbox{ if } b(i) = 2
\end{align*}
Thus, modeling the spectral embedding of the adjacency matrix as a mixture of Gaussians is not only analytically convenient but also theoretically sound. 
\end{comment}

\subsection{Pattern Recognition}

Classification tasks require labeling objects whose group membership is unknown. Generally, we consider a classifier as a function from an input space $ \mathcal{X} $ to a set of labels. Namely, $ g: \mathcal{X} \rightarrow [K] $. In the current setting we consider $ \mathbb{R} $ and $ \mathbb{R}^{d}$ as input spaces. For objects in $ \mathbb{R}^{d} $, it is intuitively appealing to think of the entire space as "painted" by $ K \in \mathbb{N} $ colors, with $ X \in \mathcal{X} $ colored $ k $ if $ g(X) = k $. $ g(\cdot) $ is typically unknown and there are numerous methods for estimating it. This paper focuses exclusively on a Bayes' plug-in classifier. 

More formally, let $ (X, Y) \sim F_{XY} $ and consider a series of observations $ \mathcal{T}_{n} = \{(X_{1}, Y_{1}), \hdots, (X_{n}, Y_{n})\} $, where $ X_{i} \in \mathcal{X} $ and $ Y_{i} \in [K] $. The goal of pattern recognition is to construct a  function from $ \mathcal{X} $ to $ [K] $  based on $ \mathcal{T}_{n} $, denoted $ g(\cdot | \mathcal{T}_{n}) $ or $ g_{n}(\cdot) $, that minimizes average loss. 

\begin{comment}
In this paper, we use the 0-1 loss function where $ \ell(a, b) = 0 $ if $ a = b $ and $\ell(a, b) = 1 $ otherwise for $ a, b \in \mathcal{Y} $. Let $ L(g_{n}(\cdot)) $ denote the average loss incurred by $ g_{n}(\cdot) $.
\end{comment}

It is well known that if the conditional density of $ X $ given $ Y $ is known then the classifier that minimizes the average 0-1 loss is given by 
% TODO: notation is strange.  what is the semicolor theta_u doing? it has not yet been introduced as a parameter
$$ g^{*}(X) = \argmax_{u \in [K]} f_{X |Y = u}(X; \theta_{u} | Y = u) P(Y = u) $$ and is called Bayes' classifier, with $ L(g^{*}(\cdot)) = L^{*} $ known as Bayes' loss. The quantities $ f_{X |Y = u}(X; \theta_{u} | Y = u) $ and $ P(Y = u) $ are typically unknown and must be estimated.

If $ \theta \in \Theta $ is estimated by $ \hat{\theta} $ and $ f(\cdot; \theta) $ is estimated by $ f(\cdot; \hat{\theta}) $ then the classifier that uses the estimated quantity is called a plug-in classifier.

See \citep{fishkind2015vertex} for pattern recognition methods for unweighted networks, \citep{chen2013robust} for a robust vertex classification method in a set of contamination settings, and \citep{DBLP:journals/corr/abs-1101-3291} for non-spectral vertex classification techniques on large networks.

\subsubsection{Univariate Normal, two class Bayes classifier}

Consider a two-class classification problem in $ \mathbb{R} $ where the generative distributions are known to be Gaussian. Furthermore, suppose that the means and variances of the two distributions are known and that $ \sigma_{1} \neq \sigma_{2} $. WOLOG suppose $ \mu_{1} < \mu_{2} $. Then the Bayes decision boundary is given by the roots of a quadratic equation (if $ \sigma_{1} = \sigma_{2} $ then the optimal decision boundary is given by a line \citep{devroye2013probabilistic}), denoted $ x_{\pm}^{*} $:
$$ {\scriptstyle x_{\pm}^{*} = \frac{\mu_{2}\sigma_{1}^{2} - \mu_{1}\sigma_{2}^{2} \pm \sqrt{\big(\mu_{1}\sigma_{2}^{2} - \mu_{2}\sigma_{1}^{2}\big)^2 - \big(\sigma_{1}^{2} - \sigma_{2}^{2}\big)\big(\mu_{2}^{2}\sigma_{1}^{2} - \mu_{1}^{2}\sigma_{2}^{2} + 2\sigma_{1}^{2}\sigma_{2}^{2}\log(\frac{\pi_{1}\sigma_{2}}{\pi_{2}\sigma_{1}})\big)}}{\sigma_{1}^{2} - \sigma_{2}^{2}}  } $$
We highlight the uni-variate case because our results are generated with a rank one SBM and so the spectral embedding of the adjacency matrix is univariate. The explicit form of the decision boundary above is simply to build an intuition as to what the classifier we propose is actually doing. Moreover, by understanding where the decision boundaries come from we can shift them by tuning parameters.

\subsection{A small example (part 1)}
Consider $ C $, the weighted, hollow and symmetric adjacency matrix that is generated from an SBM with parameters $ n = 10 $ and $ B = \begin{bmatrix} (0.8)^{2} & (0.8)(0.6) \\ (0.8)(0.6) & (0.6)^{2} \end{bmatrix} $. Suppose we know the class memberships of nodes 1, 2, 6 and 7. Namely, $ b(1) = b(3) = 1 $ and $ b(8) = b(10) = 2 $. 
$$ C = \begin{bmatrix} 0 & 2 & 2 & 0 & 2 & 2 & 0 & 0 & 0 & 1 \\
   2 & 0 & 2 & 0 & 0 & 0 & 0 & 0 & 0 & 0 \\
   2 & 2 & 0 & 2 & 0 & 0 & 3 & 4 & 0 & 4 \\
   0 & 0 & 2 & 0 & 1 & 3 & 2 & 0 & 0 & 0 \\
   2 & 0 & 0 & 1 & 0 & 0 & 3 & 0 & 0 & 0 \\
   2 & 0 & 0 & 3 & 0 & 0 & 0 & 0 & 0 & 0 \\
   0 & 0 & 3 & 2 & 3 & 0 & 0 & 0 & 4 & 0 \\
   0 & 0 & 4 & 0 & 0 & 0 & 0 & 0 & 6 & 2 \\
   0 & 0 & 0 & 0 & 0 & 0 & 4 & 6 & 0 & 0 \\
   1 & 0 & 4 & 0 & 0 & 0 & 0 & 2 & 0 & 0
\end{bmatrix} $$ \begin{figure}[!t]
\centering
\includegraphics[width=89mm]{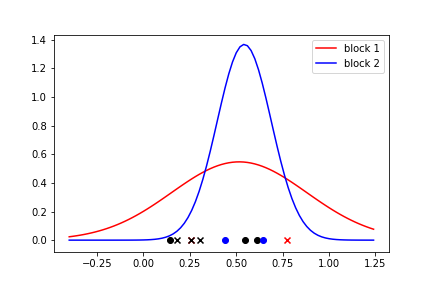}
\caption{An illustration of how to use pass-to-ranks for a classification task. The blue curve is the estimated Gaussian for block 1 and the red curve is the estimated Gaussian for block 2. The nodes from block 1 are x's and the nodes from block 2 are o's. Unlabeled nodes are black.}
\label{small_example_1}
\end{figure}

If we want to apply pass-to-ranks to $ C $ we first count the number edges (17 -- our network is undirected) and give each nonzero edge weight a rank. For the sake of clarity we will consider $ ptr(C): \mathbb{R}^{n \times n} \rightarrow [0, 1]^{n \times n} $. There is one 6, so we give it rank $ |E| = 17 $. There are three 4s, so we give them rank $ \frac{3 |E| - (1 + 2 + 3)}{3} = 15 $, and so on. Resulting in
$$ C_{ptr} = \frac{1}{17} \begin{bmatrix} 0 & \frac{13}{2} & \frac{13}{2} & 0 & \frac{13}{2} & \frac{13}{2} & 0 & 0 & 0 & \frac{3}{2} \\
\frac{13}{2} & 0 & \frac{13}{2} & 0 & 0 & 0 & 0 & 0 & 0 & 0 \\
\frac{13}{2} & \frac{13}{2} & 0 & \frac{13}{2} & 0 & 0 & 12 & 15 & 0 & 15 \\
0 & 0 & \frac{13}{2} & 0 & \frac{3}{2} & 12 & \frac{13}{2} & 0 & 0 & 0 \\
\frac{13}{2} & 0 & 0 & \frac{3}{2} & 0 & 0 & 12 & 0 & 0 & 0 \\
\frac{13}{2} & 0 & 0 & 12 & 0 & 0 & 0 & 0 & 0 & 0 \\
0 & 0 & 12 & \frac{13}{2} & 12 & 0 & 0 & 0 & 15 & 0 \\
0 & 0 & 15 & 0 & 0 & 0 & 0 & 0 & 1 & \frac{13}{2} \\
0 & 0 & 0 & 0 & 0 & 0 & 15 & 1 & 0 & 0 \\
\frac{3}{2} & 0 & 15 & 0 & 0 & 0 & 0 & \frac{13}{2} & 0 & 0
\end{bmatrix} $$ We then find the Singular Value Decomposition of $ C_{ptr} = U \Sigma U^{T} $ and take the first column of $ \hat{X} = U \Sigma^{1/2} $ as the estimated positions. That is, the latent positions can be estimated by $$ \hat{X} = [0.26, 0.18, 0.77, 0.30, 0.25, 0.14, 0.61, 0.65, 0.55, 0.44]^{T} $$ Under the assumption that the latent positions are distributed normally, we can estimate the parameters of the Gaussian mixture model, with $ \hat{\mu_{1}} = \frac{0.26 + 0.77}{2} = 0.51 $, $ \hat{\sigma_{1}} =  0.36 $, $ \hat{\mu_{2}} = \frac{0.65 + 0.44}{2} = 0.54 $, $ \hat{\sigma_{2}} = 0.15 $, resulting in the mixture in Figure \ref{small_example_1}. We will return to this example again later.

\begin{comment} Finally, we can classify an unlabeled node based on the likelihood of observing its estimated position under each Gaussian. For example, the likelihood of observing the estimated position corresponding to node 3 under block 1 is around 0.03. The likelihood under block 2 is around 0.02. Therefore, we classify node 3 as a member of block 1. 

\end{comment}

\subsubsection{A second perspective}
Consider a weighted, symmetric and hollow matrix $ C \in \mathbb{R}^{n \times n} $. Recall from section 1.1 that we can think of this matrix as the Hadamard product (denoted $ \circ $) of $ A \in \{0, 1\}^{n \times n} $ and $ W = \mathbb{R}^{n \times n} $ where $ a_{i,j} $ is 1 if there is an edge between node $ i $ and node $ j $ and 0 otherwise. $ w_{i,j} $ is the weight of the edge between node $ i $ and node $ j $. Using the $ C $ in section 2.4 as an example,

\begin{comment}
= \begin{bmatrix} 
0 & 2 & 2 & 0 & 2 & 2 & 0 & 0 & 0 & 1 \\
2 & 0 & 2 & 0 & 0 & 0 & 0 & 0 & 0 & 0 \\
2 & 2 & 0 & 2 & 0 & 0 & 3 & 4 & 0 & 4 \\
0 & 0 & 2 & 0 & 1 & 3 & 2 & 0 & 0 & 0 \\
2 & 0 & 0 & 1 & 0 & 0 & 3 & 0 & 0 & 0 \\
2 & 0 & 0 & 3 & 0 & 0 & 0 & 0 & 0 & 0 \\
0 & 0 & 3 & 2 & 3 & 0 & 0 & 0 & 4 & 0 \\
0 & 0 & 4 & 0 & 0 & 0 & 0 & 0 & 6 & 2 \\
0 & 0 & 0 & 0 & 0 & 0 & 4 & 6 & 0 & 0 \\
1 & 0 & 4 & 0 & 0 & 0 & 0 & 2 & 0 & 0 
\end{bmatrix} 
\end{comment}

\begin{align*}
C &= \begin{bmatrix} 
0 & 1 & 1 & 0 & 1 & 1 & 0 & 0 & 0 & 1 \\
1 & 0 & 1 & 0 & 0 & 0 & 0 & 0 & 0 & 0 \\
1 & 1 & 0 & 1 & 0 & 0 & 1 & 1 & 0 & 1 \\
0 & 0 & 1 & 0 & 1 & 1 & 1 & 0 & 0 & 0 \\
1 & 0 & 0 & 1 & 0 & 0 & 1 & 0 & 0 & 0 \\
1 & 0 & 0 & 1 & 0 & 0 & 0 & 0 & 0 & 0 \\
0 & 0 & 1 & 1 & 1 & 0 & 0 & 0 & 1 & 0 \\
0 & 0 & 1 & 0 & 0 & 0 & 0 & 0 & 1 & 1 \\
0 & 0 & 0 & 0 & 0 & 0 & 1 & 1 & 0 & 0 \\
1 & 0 & 1 & 0 & 0 & 0 & 0 & 1 & 0 & 0 
\end{bmatrix} \\
&\circ 
\begin{bmatrix} 
x & 2 & 2 & x & 2 & 2 & x & x & x & 1 \\ 
2 & x & 2 & x & x & x & x & x & x & x \\
2 & 2 & x & 2 & x & x & 3 & 4 & x & 4 \\
x & x & 2 & x & 1 & 3 & 2 & x & x & x \\
2 & x & x & 1 & x & x & 3 & x & x & x \\
2 & x & x & 3 & x & x & x & x & x & x \\
x & x & 3 & 2 & 3 & x & x & x & 4 & x \\
x & x & 4 & x & x & x & x & x & 6 & 2 \\
x & x & x & x & x & x & 4 & 6 & x & x \\
1 & x & 4 & x & x & x & x & 2 & x & x 
\end{bmatrix}
\end{align*}
where $ x $ is an unobserved weight between node $ i $ and node $ j $. Note that each $ x $ should be indexed by $ i $ and $ j $ but this dependence is suppressed to avoid cluttering the matrix. 

Splitting C into $ A $ and $ W $ gives us reason to consider adding another component to the SBM to address $ W $. With this in mind, it seems natural to propose a matrix of distributions, $ \mathcal{F} $, where $ \mathcal{F}_{u, v} $, $ u, v \in [K] $ is the distribution governing the edge weights between block $ u $ and block $ v $. It is clear that $ \mathcal{F} $ is analogous to the matrix $ B $ in the standard SBM, which models the adjacency relationship between nodes in block $ u $ and nodes in block $ v $. While this extension of the SBM is completely natural, the question remains about how to use this additional component for classification tasks.

% TODO: this makes me wonder, why not assume that we have a collection of K zero-inflated distributions, rather than K^2 distributions & K latent positions that collectivley are positive definite.  i feel as though your model is both over and under parameterized in different ways.

\section{Ordered Edge Weight Distributions}
Walking through the procedure in section 2.4 gives some insight as to why pass-to-ranks is an effective method. It combines the adjacency and edge weight information in a way such that neither dominates the other. However, the usefulness of the weight information still depends on there being an ordered relationship that can be captured by a simple ranking mechanism. 

Assuming that the weights of the network encode information about block membership, if we use pass-to-ranks we'd hope that the edge weights between nodes in block $ u $ and nodes in block $ v $ have some ordered relationship, i.e. $ \mathbb{E}(w_{u, v}) < \mathbb{E}(w_{u, t}) $ for $ u, v, t \in [K] $. That is, the edge weights between nodes in block $ u $ and block $ v $ come from a distribution with a different mean than the edge weights between nodes in block $ u $ and block $ t $. 
% TODO: i don't think the above is right.  if the variances are different, PTR can also work.  in fact, the only reason PTR does *not* work, at least asymptotically, is when the *scales* matter.  that is, if 0,1,4,5 is meaningfully different from 1,2,3,5.  

While we do not know order of the distributions, the partially observed $ b(\cdot) $ allows us to estimate the ordering using the weights between training data. For each unlabeled node we can estimate the ordering of its distributions using the edge weights between it and the training data. These estimated orderings can be used as proxies for $ \hat{\mathcal{F}}_{u} $ and $ \hat{\mathcal{F}}(i) $ from section 1.1.

In this section we expand on the idea of ranking objects by comparing the estimated ordering for an unlabeled node to the estimated ordering for each block. We use the results of this comparison to update the class membership priors for each unlabeled node. We demonstrate the effectiveness of this method as compared to pass-to-ranks for data that is generated from an SBM with the additional weight distribution component $ \mathcal{F} $. In an example in section 3.3 we use the footrule distance on a pair of permutations to find the dissimilarity between them. The footrule distance is the sum of absolute differences of the indices of the set of objects. That is, if we have two permutations of $ [4] $, $ P_{1} = (1, 2, 3, 4) $ and $ P_{2} = (2, 3, 4, 1) $ then $ d_{FR}(P_{1}, P_{2}) = \sum_{i = 1}^{4} |\arg_{P_{1}}(i) - \arg_{P_{2}}(i)| = |1 - 4| + |2 - 1| + |3 - 2| + |4 - 3| = 6 $, where $ \arg_{P_{u}}(i) $ returns the index of $ i $ in $ P_{j} $.
% TODO: i don't understand the footrule distance, or whether it is a proper distance, or just a dissimilarity.

\subsection{Model Assumptions}
We assume that the network is generated from a K-Block SBM with partially observed block membership function $ b(\cdot) $, unobserved $ B $ and unobserved $ \mathcal{F} $. Moreover, we assume that the edge weight distributions have finite expectation and that $ E(F_{u,v}) \neq E(F_{s,t}) $ for all $ u, v, s, t \in [K] $. It is then possible to order the distributions based on expected value, i.e. there exists an ordering such that $ E(F_{(1)}) < \hdots < E(F_{(L)}) $ where $ L = {K \choose 2} + K $ and $ F_{(i)} $ is the $ i^{th} $ ranked distribution. 
% TODO: i don't see it.  why can't they all have equal means based on the above assumptions?

Let $ O(\mathcal{F}) = (E(F_{(1)}), \hdots , E(F_{(L)})) $ be the ordering of the distributions for a specific K-block SBM with weight distribution matrix $ \mathcal{F} $ and the appropriate restrictions on $ F_{u, v} $. We sometimes refer to $ O(\mathcal{F}) $ as the "global" ordering.

The global ordering implies a collection of K "local" orderings, $ O_{u}(\mathcal{F}) = (F_{u, (1)}, \hdots, F_{u, (K)}) $ for $ u \in [K] $. Moreover, each node $ i $ in $ V $ has an associated ordering based on block membership, i.e. $\bar{O}_{i}(\mathcal{F}) = (F_{i, (1)}, \hdots, F_{i, (K)}) = O_{b(i)}(\mathcal{F}) $. We view $ O_{u} $ and $ \bar{O}_{i} $ as proxies for the vectors of estimated distributions from section 1.1. That is, instead of comparing vectors of estimated distributions directly we can compare different permutations of $ [K] $.

\subsection{Methodology}

As discussed previously and showcased in section 2.4, classification tasks for graph objects can be done via spectral methods and, in particular, using the spectral embedding of the unweighted adjacency matrix of the graph. Once the spectral embedding is obtained, any method used for Euclidian data can be applied to the estimated positions. A mixture of Gaussians is used for this method, with parameter estimations based on the training data. The partially observed block membership function, for example, can be used to estimate the block membership prior associated with each Gaussian in the mixture. That is, $ \hat{\pi} = (\hat{\pi}_1, \hdots, \hat{\pi}_K) $ where 
$$ \hat{\pi}_{u} = \frac{\sum_{i \in N_{u}} \mathbbm{1}_{\{b(i) = u\}}}{\sum_{v \in [K]} |N_{v}|} $$ for all $ u \in [K] $, where $ N_{u} $ is the set of training nodes for block $ u $.

One way to use the block membership information encoded in $ \mathcal{F} $ is to integrate it into tried and trusted procedures. There are a few things to consider when doing this. Firstly, we know that fitting a mixture of Gaussians using the spectral embedding of the unweighted adjacency matrix works well for classification tasks on unweighted networks. Secondly, our shift of perspective (section 2.4.1) and the addition of $ \mathcal{F} $ means there is more information about class membership available. Hence, unless we wish to deviate from spectral based methods, we must use the additional block membership information to update our block membership priors. 

To do this, we define a permutation error for each ordering and convert the error into a measure of similarity that is consequently used to update the prior for each unlabeled node. There are innumerable dissimilarities on permutations to consider -- footrule distance, Kendall's Tau, 0-1 error, etc. We let $ d(\cdot, \cdot) $ be the dissimilarity metric and let $ d_{i, u} $ be the dissimilarity between $ O_{u} $ and $ \bar{O}_{i} $. Namely, $ d_{i, u} = d(\bar{O}_{i}, O_{u}) $. We then define $ D_{i} = (d_{i, 1}, \hdots, d_{i, K}) $ as the error vector for unlabeled node $ i $. Normalizing the error vector results in 
$$ ND_{i} = \frac{1}{\sum_{u \in [K]} d_{i, u}} D_{i} $$ and we define the similarity vector to be
$$ S_{i} = \hat{1}_K - (ND_{i, 1}, \hdots, ND_{i, K}) $$ where $ \hat{1}_K $ is the vector of ones of length K. Finally, we update our priors
$$ \hat{\pi}_i = \frac{1}{\langle \hat{\pi}, NS_{i} \rangle} (\hat{\pi}_{1} NS_{i, 1}, \hdots, \hat{\pi}_{K} NS_{i, K}) = (\hat{\pi}_{i, 1}, \hdots, \hat{\pi}_{i, K}) $$ where $\langle\cdot,\cdot\rangle $ is the inner product of two vectors. The resulting classifier is
$$ g_{R}(x_{i}) = \argmax_{u \in [K]} \hat{\pi}_{i, u} f_{X | Y = u}(x_{i}; \hat{\theta}_{u}) $$ where $ f(\cdot; \hat{\theta}_{u} | Y = u) $ is the estimated Gaussian density for block $ u $. Notice that this new classifier utilizes both adjacency and weight information -- in short, integrated the class membership information encoded in the edge weights. 

We recognize that we reuse notation when defining the estimated class membership priors found using $ b(\cdot) $ and the updated class priors. The meaning of $ \hat{\pi}_{i} $ or $ \hat{\pi}_{u} $ should be clear in context -- one refers to the updated prior vector for node $ i $ and the other refers to the original estimated class membership prior for block $ u $.

\subsection{Properties of the updated priors classifier in the two block case}

The two block rank one case sheds light on the mechanics of the methodology. First recall the decision boundaries from section 2.3.1: 
$$ {\scriptstyle x_{\pm}^{*} = \frac{\mu_{2}\sigma_{1}^{2} - \mu_{1}\sigma_{2}^{2} \pm \sqrt{\big(\mu_{1}\sigma_{2}^{2} - \mu_{2}\sigma_{1}^{2}\big)^2 - \big(\sigma_{1}^{2} - \sigma_{2}^{2}\big)\big(\mu_{2}^{2}\sigma_{1}^{2} - \mu_{1}^{2}\sigma_{2}^{2} + 2\sigma_{1}^{2}\sigma_{2}^{2}\log(\frac{\pi_{1}\sigma_{2}}{\pi_{2}\sigma_{1}})\big)}}{\sigma_{1}^{2} - \sigma_{2}^{2}} } $$ Tuning the ratio of the block membership priors has an explicit effect on the position of the decision boundaries. Information on the ordering of the distributions allows us to move this boundary in a non-arbitrary way for each unlabeled node. 

Note that updating priors with the same error, and thus the same similarity, would result in an "update" of the priors for vertex $ i $ such that $ \hat{\pi} = \hat{\pi}_i $, i.e. the "updated" prior for vertex $ i $ would be the same as the prior estimated from the observed portion of $ b(\cdot) $ (see example below). Thus, we can focus our attention on the case where a disagreement occurs.

\graphicspath{}
\begin{figure}[t!]
\centering
\includegraphics[width=90mm]{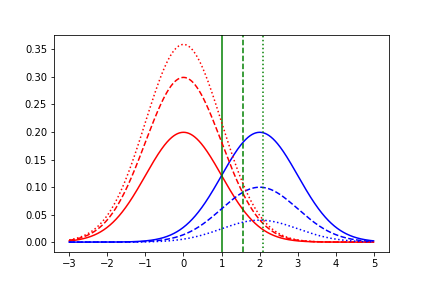}
\caption{An illustration of how changing the priors associated with each Gaussian changes the mixture densities and, hence, moves the decision boundary for $ \mu_1 = 0, \mu_2 = 2, \sigma = 1, \pi_1 = \{ 0.5 \mbox{ (solid)}, 0.75 \mbox{ (dashed)}, 0.9 \mbox{ (dotted)}$\}.}
\label{ref: how_priors_change_DB}
\end{figure}
Without loss of generality, assume $ O_{1} = (1, 2) $ and $ O_{2} = (2, 1) $. Then for an unlabeled node $ i $, $ \bar{O}_{i} $ is equal to $ O_{1} $ or $ O_{2} $. To illustrate the mechanics of the method, let $ \bar{O}_{i} = O_{1} $. When we include a base error of one, discussed in detail below, and use the footrule distance, we obtain a normalized error vector $ ND_{i} = (\frac{1}{4}, \frac{3}{4}) $ and corresponding similarity vector $ S_{i} = \hat{1}_2 - ND_{i} = (\frac{3}{4}, \frac{1}{4}) $, resulting in 
$$ \hat{\pi}_{i} = \frac{1}{\frac{3}{4}\hat{\pi}_1 + \frac{1}{4}\hat{\pi}_2} (\frac{3}{4} \hat{\pi}_1, \frac{1}{4} \hat{\pi}_2) $$ which leads to new decision boundaries specific to the particular unlabeled node, $ x^{*}_{\pm, i} $.

\begin{comment}
$$ {\scriptstyle x_{\pm, i}^{*} = \frac{\mu_{2}\sigma_{1}^{2} - \mu_{1}\sigma_{2}^{2} \pm \sqrt{\big(\mu_{1}\sigma_{2}^{2} - \mu_{2}\sigma_{1}^{2}\big)^2 - \big(\sigma_{1}^{2} - \sigma_{2}^{2}\big)\big(\mu_{2}^{2}\sigma_{1}^{2} - \mu_{1}^{2}\sigma_{2}^{2} + 2\sigma_{1}^{2}\sigma_{2}^{2}\log(\frac{\hat{\pi}_{i, 1}\sigma_{2}}{\hat{\pi}_{i, 2}\sigma_{1}})\big)}}{\sigma_{1}^{2} - \sigma_{2}^{2}} } $$
\end{comment} 

\graphicspath{}
\begin{figure}[t!]
\centering
\begin{subfigure}[b]{.49\linewidth}
\includegraphics[width=\linewidth]{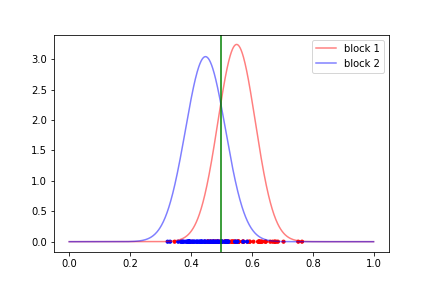}
\caption{Original mixture}\label{fig:ob_og}
\end{subfigure}

\begin{subfigure}[b]{.49\linewidth}
\includegraphics[width=\linewidth]{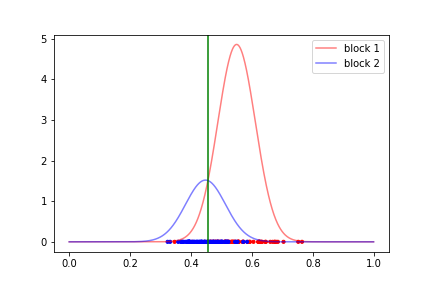}
\caption{Mixture when $ \bar{O}_{i} = O_{1} $}\label{fig:ob_1}
\end{subfigure}
\begin{subfigure}[b]{.49\linewidth}
\includegraphics[width=\linewidth]{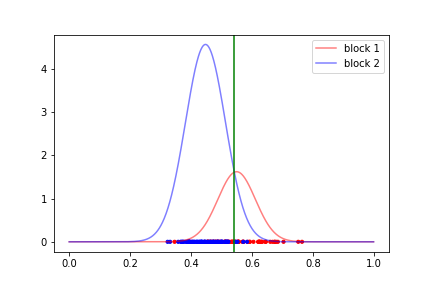}
\caption{Mixture when $ \bar{O}_{i} = O_{2} $} \label{fig:ob_2}
\end{subfigure}
\caption{a) shows the densities and corresponding decision boundary associated with the original priors. b) shows the densities and corresponding decision boundary when an unlabeled node's ordering matches with the ordering for block 1. c) shows the densities and corresponding decision boundary when an unlabeled node's ordering matches the ordering for block 2. Please note that there are actually two decision boundaries for each case (since the variances of the Gaussians are typically unequal) but we have muted the less impactful one. From b) and c) it is clear that better classification results can be achieved with an informed update.}
    
\label{fig:ordered_example}
\end{figure}
Figure \ref{fig:ordered_example} shows how updating priors changes the decision boundary for particular nodes in the setting where $ B = \begin{bmatrix} (0.55)^{2} & (0.55)(0.45) \\ (0.55)(0.45) & (0.45)^{2} \end{bmatrix} $, $ \pi_{1} = \pi_{2} = 0.5 $, $ n = 150 $, $ \mathcal{F} = \begin{bmatrix}
N(5, 1) & N(10, 1) \\ N(10, 1) & N(5, 1)
\end{bmatrix} $. From the figure we can see that an informed shift in the decision boundary can have a huge impact on classification results. For example, the right most plot in Figure \ref{fig:ordered_example} would correctly classify more unlabeled nodes whose latent block is block 2 than the original classifier. Again, moving the decision boundary in an informed way can decrease misclassification rates. Simulation results are discussed thoroughly in section 3.5.

The base error of one that we applied is called plus-one smoothing and is generally used to avoid method or model degradation, see \citep{gale1994s}. In our case, if we did not apply it we would classify unlabeled nodes solely on the information contained in the edge weights. A simple way to see this is in the example we presented above. Recall that $ O_{1} = \bar{O}_{i} = (1,2) $ and $ O_{2} = (2, 1) $. If we did not apply plus-one smoothing we would end up with the error vector $ (0, 2) $, which yields the normalized error vector $ (0, 1) $ and the similarity vector $ (1, 0) $. The updated prior would be $ (1, 0) $ and we would completely ignore all other block membership information when classifying. 

Interestingly, additive smoothing can have a significant impact on our procedure. Imagine that instead of plus-one, we used plus-10000 smoothing. Then, doing the same as before, we get the error vector $ (10000, 10002) $, the normalized error vector $ (\frac{10000}{20002}, \frac{10002}{20002}) \approx (0.5, 0.5) $ and, finally, the similarity vector that is approximately $ (0.5, 0.5) $. But this similarity vector gives us no additional class membership information! In fact with the current method, plus-10000 smoothing would spit out approximately our original priors. In section 3.6 we look at a dynamic type of additive smoothing that is less naive than plus-one smoothing and less rigid than plus-10000 smoothing.

\subsection{A small example (part 2)}
\begin{figure}[t!]
\centering
\includegraphics[width=90mm]{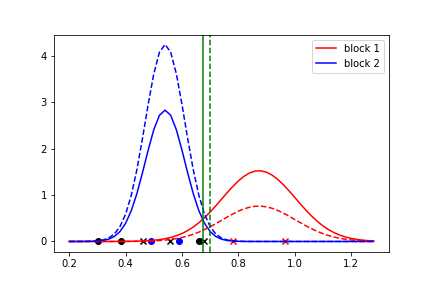}
\caption{An illustration of how to update priors for a classification task. The solid blue curve is the estimated Gaussian for block 1 and the solid red curve is the estimated Gaussian for block 2. Their dashed counterparts are the curves corresponding to the densities after the prior update for node 6. The nodes from block 1 are x's and the nodes from block 2 are o's. Unlabeled nodes are black. The green solid line is the relevant decision boundary from the prior estimated from the observed block membership function. The green dashed line is the decision boundary after we updated the prior using information from the ordering of the edge weight distributions assuming $ \bar{O}_{i} = O_{2} $. In this particular example there seems be little gained in shifting the decision boundary.}

\label{example_of_updating_priors}
\end{figure}
Consider matrix $ C $ from section 2.4.1.  This time we use the spectral embedding of the unweighted adjacency matrix to estimate the latent positions. We obtain estimated positions 

$$ \hat{X} = [0.78, 0.46, 0.97, 0.68, 0.56, 0.38, 0.66, 0.49, 0.30, 0.59]^{T} $$
Then, with $ b(1) = b(3) = 1 $ and $ b(8) = b(10) = 2 $ known, we estimate the Gaussian parameters to obtain $ \hat{\mu_{1}} = 0.62 $, $ \hat{\sigma_{1}} = 0.226 $, $ \hat{\mu_{2}} = 0.52 $, and $ \hat{\sigma_{2} = 0.198} $, resulting in the densities in Figure \ref{example_of_updating_priors}. 

To implement the newly proposed method we must first estimate the orderings for each block, which requires estimating three means. The mean of the edge weights 1) between training data within block 1; 2) between training data from block 1 and block 2; 3) between training data within block 2. In our case we get $ \bar{X}_{1,1} = 2 $, $ \bar{X}_{1,2} = \bar{X}_{2,1} = 3 $, $\bar{X}_{2,2} = 2 $ which lead to the local orderings $ \hat{O}_{1} = (1,2) $ and $ \hat{O}_{2} = (2,1) $. 

Now consider the ordering associated with node 6, $ \bar{O}_{6} = (2, 1) $. We calculate the footrule distance and add one to get $ S_{6} = (1/4, 3/4) $. The new class membership priors are then given by $ \hat{\pi}_{6, 1} = 1/4 $ and $ \hat{\pi}_{6, 2} = 3/4 $. These new priors lead to a new mixture and, hence, new decision boundaries (the dashed curves in Figure \ref{example_of_updating_priors}).

\subsection{Results from generated data}
We look at four different settings for the two block rank one SBM with Gaussian edge weight distributions. 1) Different means and different scales; 2) Different means and same scales; 3) Same mean and different scales; 4) Same mean and same scales. Notice that for settings 1) and 2) the order assumption holds because the distributions have different means. All the simulations have $ B = \begin{bmatrix} (0.52)^{2} & (0.52)(0.48) \\ (0.52)(0.48) & (0.48)^{2} \end{bmatrix} $, $ \mathcal{F} = \begin{bmatrix}
N(\mu_{1}, \sigma_{1}^{2}) & N(\mu_{2}, \sigma_{2}^{2}) \\ N(\mu_{2}, \sigma_{2}^{2}) & N(\mu_{1}, \sigma_{1}^{2}) \end{bmatrix}$, and $ n \in [150, 200, 250, 300, 350, 400, 450, 500] $ where the number of training data is $ \frac{n}{10} $ with $ \pi_{1} = \pi_{2} = 0.5 $. 

\begin{figure}[t!]
    \centering
    %% Begin set of four figures
    \begin{subfigure}[b]{.47\linewidth}
        \includegraphics[width=\linewidth]{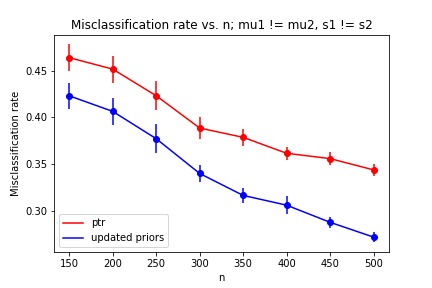}
    \end{subfigure}
    \begin{subfigure}[b]{.47\linewidth}
        \includegraphics[width=\linewidth]{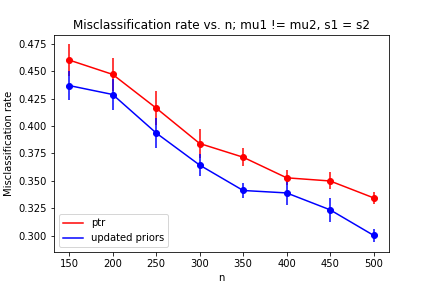}
    \end{subfigure}
    \begin{subfigure}[b]{.47\linewidth}
        \includegraphics[width=\linewidth]{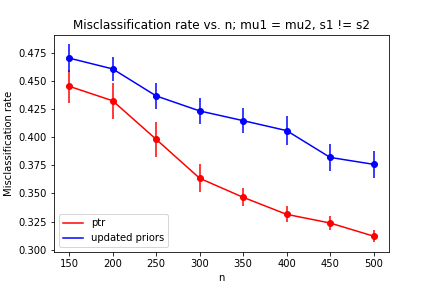}
    \end{subfigure}
    \begin{subfigure}[b]{.47\linewidth}
        \includegraphics[width=\linewidth]{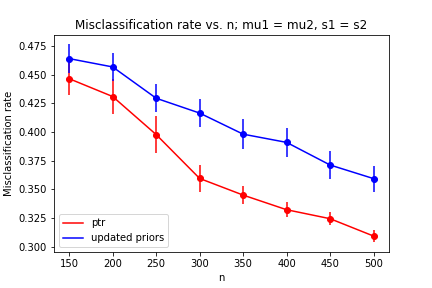}
    \end{subfigure}
    \caption{The figure on the top left plot show the results for setting 1. The top right figure show the results for setting 2. The figure on the bottom left is for setting 3 and the figure on the bottom right is for setting 4. See section 3.5 for analysis.}
    \label{fig:ordered_results}
\end{figure}

For settings 1) and 2) $ \mu_{2} - \mu_{1} = 2 $. For settings with equal variances, $ \sigma_{1} = \sigma_{2} = 9 $. When they are not equal, $ \sigma_{1} = 4 $ and $ \sigma_{2} = 9 $. Networks are generated conditioned on the number of nodes and training data in each block. Figure \ref{fig:ordered_results} shows the misclassification rate versus the number of nodes in the network. Error bars represent the 95\% confidence interval for the average of 100 iterations.

In the top two plots of Figure \ref{fig:ordered_results}, both the proposed classifier (referred to as updated priors) and quadratic discriminant analysis following the embedding of the transformed weighted adjacency matrix (referred to as pass-to-ranks) tend to perform better with a larger node set. This is reassuring and can be attributed to the fact that the adjacency spectral embedding is at the core of both methods. Another reason for the similar trends in settings 1) and 2) is that pass-to-ranks and updated priors use the edge weight information in a similar way when the means are actually different. This is especially true when the variances are the same. In fact, the difference between the two plots can be attributed to the variances being equal in one setting, which pass-to-ranks can naturally take advantage of, and the variances being different in the other.

For the bottom two plots of Figure \ref{fig:ordered_results}, the results are essentially flipped -- pass-to-ranks outperforms updated priors. This is likely due to the fact that, while pass-to-ranks does not ignore the edge weights, it does not attempt to use them in any explicit manner to determine class membership. In other words, when the edge weights do not encode information, or the method is ill-equipped to use it, any attempt to explicitly use this non-information costs a lot in terms of misclassification. A few ways to address this issue are discussed in section 3.6.

We also consider edge weights that were generated from Poisson distributions. In particular, we consider the weight distribution matrix $ \mathcal{F} = \begin{bmatrix} Pois(\mu_{1}) & Pois(\mu_{2}) \\ Pois(\mu_{2}) & Pois(\mu_{1}) \end{bmatrix} $ with the same $ n $ and $ B $ as before. We ran each simulation 100 times. The case where the order assumption does not hold again leaves some room for improvement.

\begin{figure}[t!]
    \centering
    \begin{subfigure}[b]{.49\linewidth}
    \includegraphics[width=\linewidth]{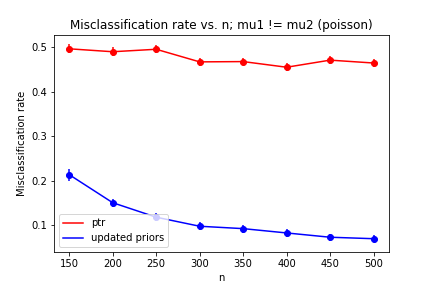}
    \end{subfigure}
    \begin{subfigure}[b]{.49\linewidth}
    \includegraphics[width=\linewidth]{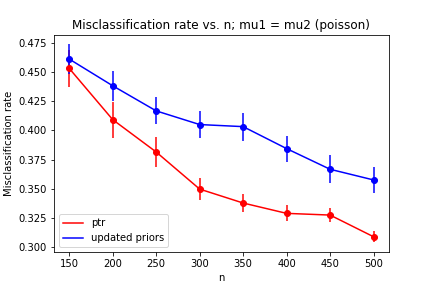}
    \end{subfigure}
    \caption{The figure on the left was generated with $ \mu_{1} = 3 $ and $ \mu_{2} = 6 $. The right figure was generated with $ \mu_{1} = \mu_{2} = 3 $. Clearly, updating priors when the order assumption holds results in better classification results as compared to pass-to-ranks. The opposite is true when the assumption does not hold.}
\end{figure}

\subsection{Testing for a Difference in the Means}

As we see in the results presented in section 3.5, the proposed classifier performs  extremely well in classification tasks when the order assumption holds. The same can not be said when the assumption fails. For this method to be robust to model misspecification, it is necessary to check if the ordering assumption holds before proceeding to update the priors. Hence, we check the assumption via hypothesis testing. We consider the null $ E(F_{u, v}) = E(F_{s, t}) $ for all $ u, v, s, t \in [K] $ against the alternative $ E(F_{u, v}) \neq E(F_{s,t}) $ for any $ u, v, s, t \in [K]$. We continue to focus on the two block case.

\begin{figure}[t!]
    \centering
    \includegraphics[width=89mm]{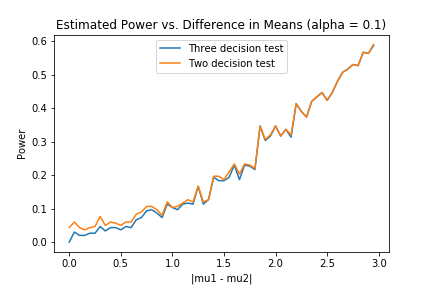}
    \caption{Estimated power curves the three decision test and two decision test for an SBM with $ n = 200 $, $ \pi_{1} = \pi_{2} = 0.5 $, number of seeds $ = \frac{n}{10} $, $ B = \protect\begin{bmatrix}
    (0.52)^2 & (0.52)(0.48) \\ (0.52)(0.48) & (0.48)^{2}
    \protect\end{bmatrix} $}
    % , $ \mathcal{F} = \begin{bmatrix}
    % N(\mu_{1}, 16) & N(\mu_{2}, 16) \\ N(\mu_{2}, 16) & N(\mu_{1}, 16) \end{bmatrix} $.} 
    % The figure was generated from 300 simulation runs for $ \mu_{2} = (\mu_{1} - 3, \mu_{1} + 3) $ with resolution $ 0.05 $.}
    \label{fig:power_curves}
\end{figure}

Here we also care about which ordering holds, i.e. $ E(F_{1, 2}) < E(F_{1,1}) $ or $ E(F_{1,1}) < E(F_{1,2}) $. We are in a testing situation where our action can take on three values. We can fail to reject the null, we can reject null and decide $ E(F_{1, 2}) < E(F_{1,1}) $, or we can reject the null and decide $ E(F_{1, 1}) < E(F_{1,2}) $. To perform this test in our setting we need a non-parametric test like the Mann-Whitney U (MWU) test, which tests for the equality of the locations of the distributions.

First, we calculate the p-value associated with the test statistic. If the p-value is less than some pre-selected $ \alpha $ then we reject the null. Then, if $ E(F_{1,1}) < E(F_{1,2}) $ we decide that $ E(F_{1,1}) < E(F_{1,2}) $. Otherwise we decide that $ E(F_{1,2}) < E(F_{1,1}) $. If we choose $ \alpha $ to be large then we are more likely to reject the null and proceed to update the priors. Here the choice of $ \alpha $ can reflect our willingness to move the decision boundaries for each node.

If we'd like to discuss how a test behaves under the null and under the two alternatives, we must first define errors in this testing scenario and, subsequently, define power. There are three types of error associated with the proposed test. Type I error, which is to incorrectly reject the null; Type 2 error, which is to incorrectly fail to reject the null; and Type 3 error, which is to correctly reject the null but incorrectly assign the order. We define power to be the probability of correctly rejecting the null and correctly ordering the distributions. 

We resort to simulation to gain insight on the properties of this test in our setting. Figure \ref{fig:power_curves} gives the power curves for the three decision test, along with the two decision test for reference. The complete simulation setting is given in the caption under the figure. It is important to point out that the three decision test has less power for $ \mu_{2} $ close to $ \mu_{1} $ but, as the difference $ |\mu_{1} - \mu_{2} | $ increases, the power curves are indistinguishable. We also note that the plot is symmetric about $ \mu_{1} - \mu_{2} = 0 $ due to the equal scale setting. While we do not correctly reject often in the settings we consider in section 3.5 (where $ |\mu_{1} - \mu_{2}| = 2) $ for $ \alpha = 0.1 $, the selection of $ \alpha $ is arbitrary and so it is unclear how we should interpret these results.

\begin{figure}[t!]
    \centering
    \begin{subfigure}[b]{.49\linewidth}
    \includegraphics[width=\linewidth]{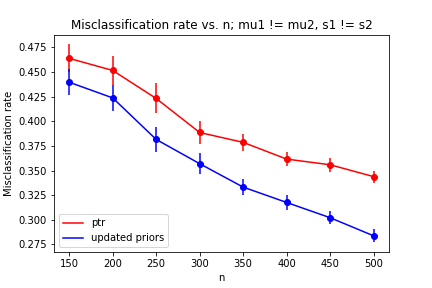}
    \end{subfigure}
    \begin{subfigure}[b]{.49\linewidth}
    \includegraphics[width=\linewidth]{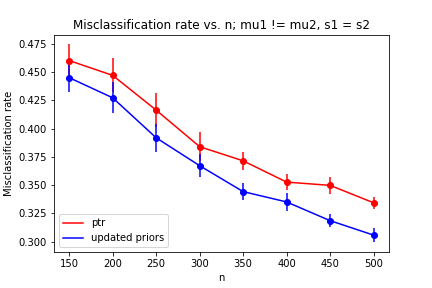}
    \end{subfigure}
    \begin{subfigure}[b]{.49\linewidth}
    \includegraphics[width=\linewidth]{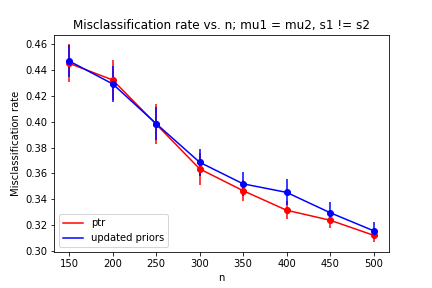}
    \end{subfigure}
    \begin{subfigure}[b]{.49\linewidth}
    \includegraphics[width=\linewidth]{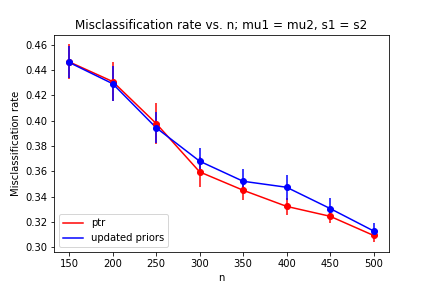}
    \end{subfigure}
    \caption{Simulation settings revisited where a hypothesis test (alpha = 0.1) for the difference in the means used to determine if we update priors. If we fail to reject we classify using the adjacency spectral embedding of the unweighted network. We can see that the testing procedure makes our procedure a bit more robust. The two charts are very similar since we fail to reject often.}
    \label{different_means_TEST}
\end{figure}

Incorporating the results from the test into the proposed method is simple: Update the priors if we reject the null and keep the original priors otherwise.

In Figure \ref{different_means_TEST} we revisit the simulation settings from before and now incorporate a hypothesis test for a difference in the means. We see from the top two plots in Figure  \ref{different_means_TEST} that our method is still preferred over pass to ranks when the order assumption holds. In the settings where the order assumption does not hold, our method is outperformed but the gap between the two methods is smaller.

\subsubsection{Dynamic Additive Smoothing}

We can also use the output of the test to inform the additive smoothing by changing the plus one smoothing to plus $ q(\cdot) $ smoothing, where $ q: [0, 1] \rightarrow [1, r] $. This can be thought of as taking a $ p $ value as an input and outputting a real number between 1 and $ r $, where $ r \in \mathbb{R} $ is "large". In our setting we first have to apply a function to a collection of $ p $ values to give us a single value in $ [0, 1] $. In the simulation study we use Fisher's Method (see section 4.1) to combine $ p $ values. Recall that if we were to use plus $ r $ smoothing then we would essentially not update our priors (see section 3.3). 

\begin{figure}[t!]
    \centering
    \begin{subfigure}[b]{.49\linewidth}
    \includegraphics[width=\linewidth]{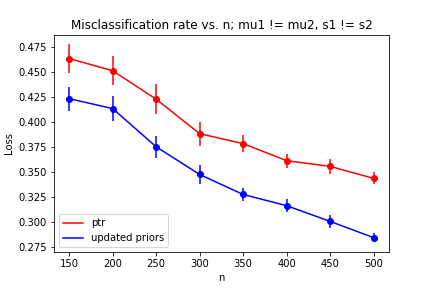}
    \end{subfigure}
    \begin{subfigure}[b]{.49\linewidth}
    \includegraphics[width=\linewidth]{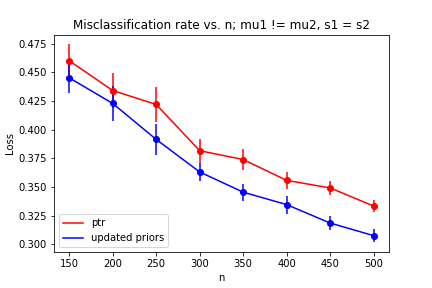}
    \end{subfigure}
    \begin{subfigure}[b]{.49\linewidth}
    \includegraphics[width=\linewidth]{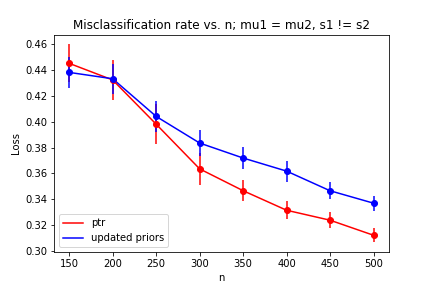}
    \end{subfigure}
    \begin{subfigure}[b]{.49\linewidth}
    \includegraphics[width=\linewidth]{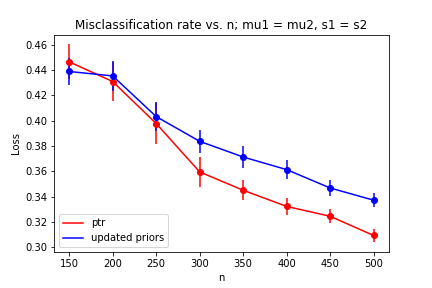}
    \end{subfigure}
    \caption{Simulation using dynamic additive smoothing to create a more robust classification procedure for settings 1-4.}
    \label{fig:DAS}
\end{figure}

Here we are just using the fact that we can interpret a small $ p $ value as evidence against the null. We consequently inform our additive smoothing procedure instead of operating on a binary test result. We can use additive smoothing to put us in a space that is operationally between the null and the alternative. 

Figure \ref{fig:DAS} shows simulation results for dynamic additive smoothing, with a story similar to the results of Figure \ref{different_means_TEST}. One important distinction, however, is that the performance of pass-to-ranks and updated priors are a bit more separated in settings 1) and 2). Using dynamic additive smoothing results in improved performances for settings 3 and 4.

Dynamic additive smoothing is just one way to use a p-value to generate a more robust (or sensitive) procedure to edge weight noise. For example, to emphasize the results of the testing procedure one could make the result more "extreme" by using a function of the p-value as the similarity metric subsequently used to update priors. In the real data analysis in Section 5, we apply a logit function with varying coeefficients to a collection of p-values to tune the method's sensitivity to edge weights.

\section{General Edge Weight Distributions}

In this section we modify the assumptions on the edge weight distributions but continue to use a measure of similarity to update priors. The methods that are proposed here are similar in spirit to the one proposed in section 3 -- simply replace the $ S_{i} $ of section 3 with the $ S_{i} $ of this section to obtain updated class membership priors to use for classification.

In this section we treat the most general edge weight distribution matrix that is brought up in section 1.1, and is the motivating setting for the majority of the preceding analysis. Recall that here we are going to deal directly with the empirical cumulative distributions. We compare vectors of empirical cumulative distributions for each block and to the corresponding vector for each unlabeled node. Luckily for us, we do not need to invent the wheel for these types of comparisons and can, instead, use a transformation of the p-values from a collection of Kolmogorov-Smirnov (KS) 2-sample tests \citep{d1975nonparametrics} to obtain a measure of similarity and subsequently update our class membership priors. 

Fisher's Method is one way to transform a collection of $ p $ values into a single $ p $ value. The method uses the fact that $ T = -2 \sum_{i= 1}^{K} p_{i} \sim \chi^{2}_{2K} $. This follows from applying the inverse transform method to a random variable distributed exponential(1) and then scaling it by a factor of two to obtain a $\chi^{2} $ distribution with 2 degrees of freedom. Finding the p value associated with the collection of $ p $ values then comes down to calculating the "extremeness" of Fisher's $ T $.

\subsection{Methodology}
\begin{figure}[t!]
    \centering
    \includegraphics[scale = 0.5]{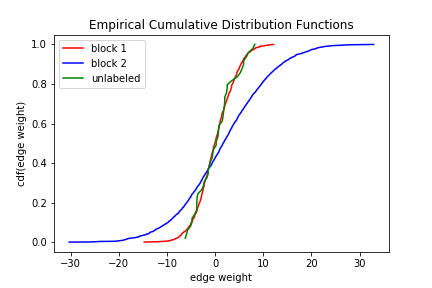}
    \caption{An illustration of how empirical cumulative distributions can encode class membership information.}
    \label{fig:ecdf_example2}
\end{figure}

We first re-introduce the notation in section 1.1. That is, we denote $ \mathcal{F}_{u} $ as the vector of empirical cumulative distribution functions corresponding to block $ u $ and $ \mathcal{F}(i) $ as the vector of empirical cumulative distribution functions corresponding to the unlabeled node $ i $. Figure \ref{fig:ecdf_example2} gives some intuition into what we are looking for when we are define a similarity metric on the space of empirical distribution functions. If we were classifying solely on the information in Figure \ref{fig:ecdf_example2} we'd clearly label the unlabeled node as block 1. Of course, this is not the only class membership information available, so we should convert this intuition into a similarity metric and then update our priors as before. 

The two-sample Kolmogorov-Smirnov test, which tests $ F_{1} = F_{2} $ against $ F_{1} \neq F_{2} $ yields a p-value that can be interpreted as a similarity metric. To make this clear, we need some notation. Let $ \mathcal{F}(i)_{v}$ be the distribution governing the edge weights between unlabeled node $ i $ and block $ v $. Similarly, let $ \mathcal{F}_{u, v}$ be the distribution governing the edge weights between block $ u $ and block $ v $. Since our unlabeled node is from one of the $ K $ blocks, this means that $ \mathcal{F}(i)_{v} = \mathcal{F}_{u, v} $ for some $ u $. Then a natural test to perform is $ \hat{\mathcal{F}}(i)_{v} = \mathcal{F}_{u, v} $ against the two-sided alternative for all $ u $. The p-value from this test can then be used as a building block for a similarity metric on this space. Holding $ u $ constant and performing this test across all $ v $ we get a collection of p values corresponding to block $ u $. Then, combining the p-values can be done using Fisher's method, $ T_{i, u} = -2 \sum_{j = 1}^{K} \log(p_{i, u, j}) \sim \chi^{2}_{2K} $ where $ p_{i, u, j} $ is the p value resulting from the test $ \hat{\mathcal{F}}(i)_{j} = \hat{\mathcal{F}}_{u, j} $. We denote the $ p $ value associated with $ T_{i,u} $ as $ p_{i, u} $. If we let $ S_{i} = (p_{i, 1}, \hdots, p_{i, K}) $ then updating priors is as before, i.e.
$$ \hat{\pi}_{i} = \frac{1}{\langle \pi, S_{i} \rangle}(\pi_{1} p_{i, 1}, \hdots, \pi_{K}p_{i, K}) $$ and the resulting classifier is
$$ g_{G}(i) = \argmax_{u \in [K]} \hat{\pi}_{i, u} f_{j}(x_{i} | b(i) = u) $$ where $ G $ is homage to the general treatment of the edge weight distributions.

\subsection{Results from generated data}
\begin{figure}[t!]{}
    \centering
    \begin{subfigure}[b]{.49\linewidth}
    \includegraphics[width=\linewidth]{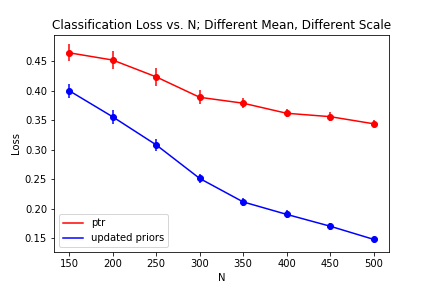}
    \end{subfigure}
    \begin{subfigure}[b]{.49\linewidth}
    \includegraphics[width=\linewidth]{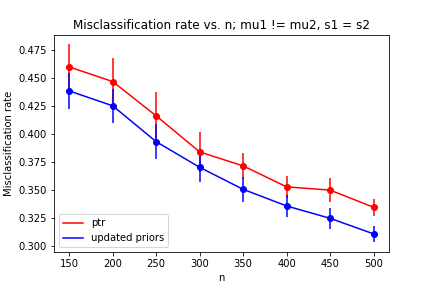}
    \end{subfigure}
    \begin{subfigure}[b]{.49\linewidth}
    \includegraphics[width=\linewidth]{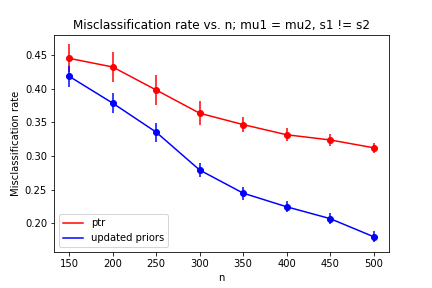}
    \end{subfigure}
    \begin{subfigure}[b]{.49\linewidth}
    \includegraphics[width=\linewidth]{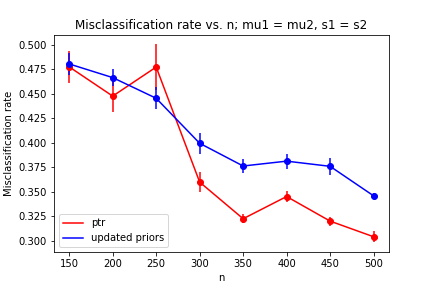}
    \end{subfigure}
    \caption{Simulations with a general edge weight distribution assumption. Similarities are based on a series of Kolmogorov-Smirnov p-values. This approach allows for the utilization of a difference in scales as well as a difference in means -- hence the improvement in setting 3.}
    \label{fig:ecdf}
\end{figure}

For our simulation study we return to the settings in section 3. The top two plots of Figure \ref{fig:ecdf} show the effectiveness of our proposed classifier for settings 1) and 2), which corresponds to settings where $ \mu_{1} \neq \mu_{2} $. In fact, we do not lose much compared to the order assumptions even when the scales are the same -- which is the setting we'd expect the classifier built on the order assumption to do better. Our new classifier, however, clearly outperforms $ g_{R}(\cdot) $ in setting 2). This is attributable to the fact that the KS test is able to account for a difference in scale and a difference in means. 

The bottom two plots of Figure \ref{fig:ecdf} look at settings 3) and 4), or the settings where the order assumption does not hold. We see, on the left, that $ g_{G}(\cdot) $ is able to outperform pass-to-ranks by accounting for scale. When there is no information in the edge weights pass-to-ranks still outperforms our classifier. 

It has become clear that we are able to leverage class membership information encoded in the edge weights to create better classifiers when the edge weights actually encode class membership information. In setting 4, pass-to-ranks will continue to outperform any classifier that makes explicit assumptions on the edge weights simply because we introduce more variance into our model. It is possible to mitigate the effect of misspecification by considering the edge weights in the discussion below.

\section{C. elegans connectome}

\begin{figure}[t!]
    \centering
    \includegraphics[scale = 0.5]{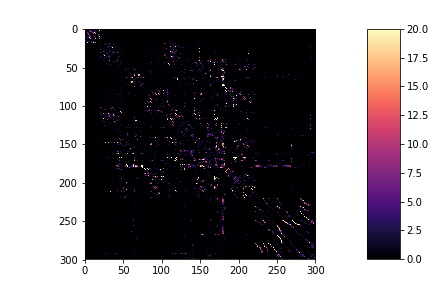}
    \caption{The C. elegans hermaphrodite weighted and symmetric connetome.}
    \label{fig:celegans_connectome}
\end{figure}

In this section we apply the classifier presented in Setion 4 to a biological data set. In particular, we consider an induced subgraph of the weighted and directed C. elegans connectome \citep{hall1991posterior} and classify an unlabled neuron as a motor, sensory or interneuron. To use the above classifier "out of the box" we symmetrize the network by taking the sum of the edge weights in the directed graph. Figure \ref{fig:celegans_connectome} shows the network where every edge weight greater than or equal to 20 is given the value 20 for visualization purposes.

\begin{figure}[t!]
    \centering
    \includegraphics[scale = 0.5]{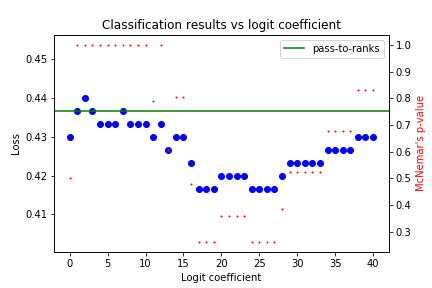}
    \caption{Classification results using updated priors for various logit coefficients (with the logit functions centered at 0.5). Large coefficients can be interpreted as being "sensitive" to the edge weight distributions in that logit functions with large coefficients send values greater than 0.5 closer to 1 and values less than 0.5 closer to 0, which can greatly affect the final "likelihoods" used for classification. The right axis shows the p-value from McNemar's test for the classifier for each logit coefficient vs pass-to-ranks.}
    \label{fig:analysis}
\end{figure}

It is important to recall and contextualize the assumptions underlying the model for which the spectral embedding is the "right" thing to do. That is, recall the assumption that the probability that a connection between two neurons exists is a function only of the type of the two neurons. This assumptions is not unreasonable -- it is only to posit that motor neurons are more likely to be connected to other motor neurons or interneurons than to sensory neurons. Furthermore, the assumptions placed on the edge weights imply that the strength of the connection is conditionally independent of the existence of an edge.

Biological implications aside, Figure \ref{fig:analysis} shows that updated priors outperforms pass-to-ranks for the majority of choices of a logit coefficient. This likely means that the updated priors classifier is more effective at using the class membership information encoded in the edge weights for discriminant analysis. The difference in classification results for different logit coefficients leaves room for model selection procedure, though we do not pursue that here.

\section{Discussion}

The preceding analysis is an introduction to the types of methods that can be used for node classification on weighted networks when it is assumed that the adjacency and edge weight information are conditionally independent. We showed that this class of methods can improve results for classification, as compared to pass-to-ranks, when the edge weights encode class membership information.

While the methods above are effective when there is class membership information encoded in $ \mathcal{F} $, we do not address all assumptions on $ \mathcal{F} $.

One class of assumptions not treated here is the set of parametric assumptions. The main benefit of parametric methods in this setting is the ability to use likelihoods as a measure of similarity. Consider the case where the edge weights do not encode any class membership information (i.e. simulation setting 4). As $ n $ gets large, the plug-in distributions will converge to the true distributions. This means that if two distributions are actually the same (i.e. $ F_{1,2} = F_{2,2}) $ the likelihood of observing the edge weights for an unlabeled node will be approximately equivalent under the two estimated distributions. When we update the priors there will be but a small change, reflecting the similarity of the distributions. Thus, the parametric framework is more flexible than the ordering assumption presented in section 3.

An interesting approach to solve the issue of misspecification (i.e. setting 4) is to use a model selection procedure to estimate the number of unique edge weight distributions. We consider this as an alternative (and perhaps more direct) method to the "plus q(p)" smoothing presented above.

We do not claim that this class of methods is the most effective way to use this information. We also make no claim as to how these methods would perform if the parameters governing $ B $ and $ \mathcal{F} $ are related in any way. It is unclear if we would even want to stay in the spectral embedding framework. 

We would also like to point out that the methodology used in this paper is not limited to a weighted network setting. Current research is being conducted in when exactly a classifier specific to the testing point is useful. On a similar note, our focus in this paper is on the supervised setting. Extensions to the unsupervised setting is natural and is currently being investigated.

\begin{comment}
\section{Conclusion}
In particular, in section 3 we proposed an effective method for node classification when $ \mathcal{F} $ can be ordered. We also presented different ways to deal with model misspecification.

In section 4 we treated general edge weight distributions and showed that the only setting in which pass-to-ranks is preferred is when the the edge weight distributions do not encode class membership.

\end{comment}

\bibliographystyle{IEEEtran}

% \nocite{*}
% \bibliography{citations}
% \bibliographystyle{plainnat}
\bibliography{main}

% \IEEEtriggeratref{8}






% \end{thebibliography}


\begin{IEEEbiography}[{\includegraphics[width=1in,height=1.25in,clip,keepaspectratio]{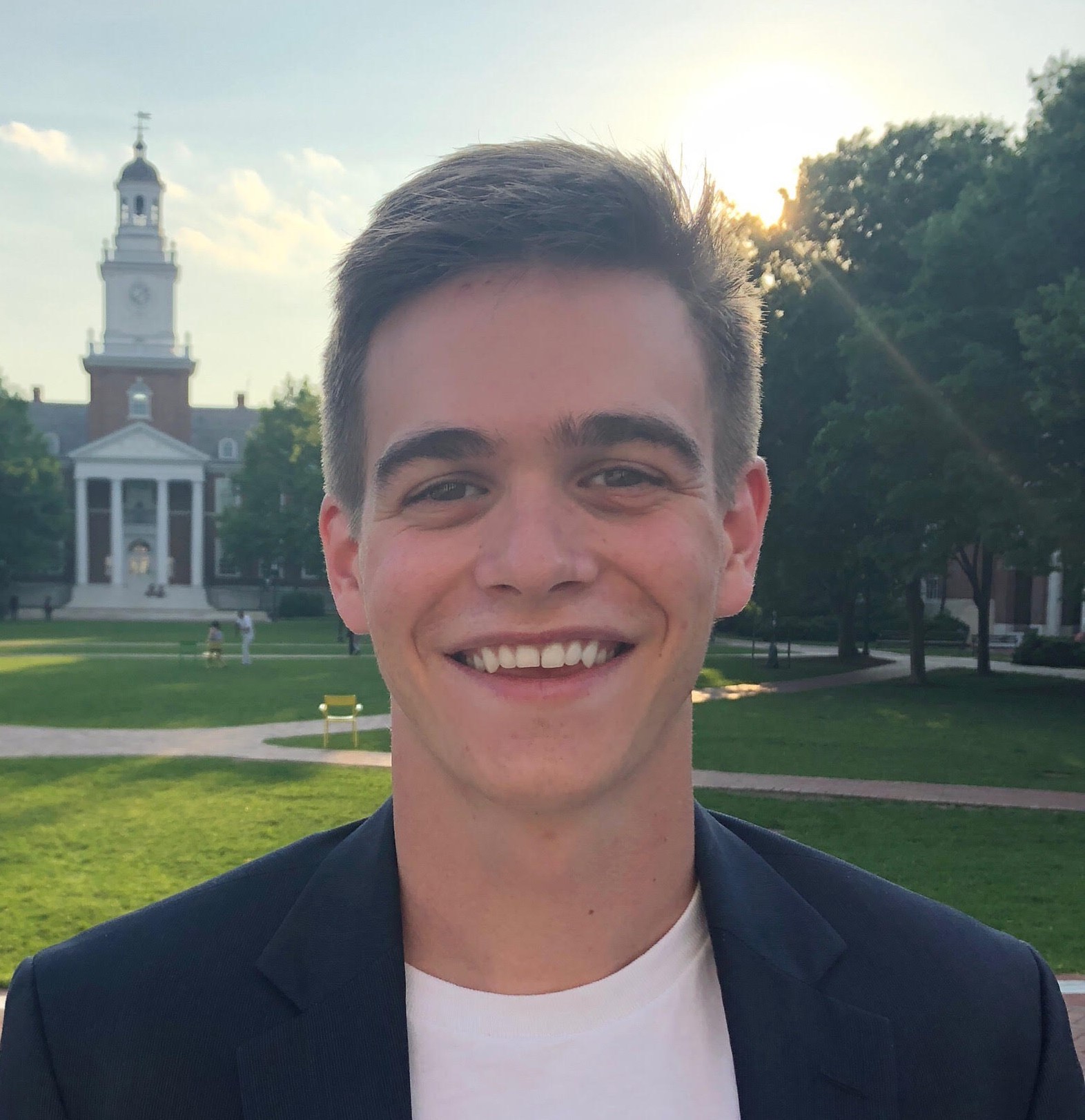}}]{Hayden Helm}
received the BS and the MSE degree in Applied Math and Statistics with focuses in Statistics and Statistical Learning from Johns Hopkins University in 2018. He currently works as an Assistant Research Engineer at the Center for Imaging Sciences at Johns Hopkins University. His current research interests lie in the intersection of statistical pattern recognition and network analysis. 
\end{IEEEbiography}

\begin{IEEEbiography}[{\includegraphics[width=1in,height=2in,clip,keepaspectratio]{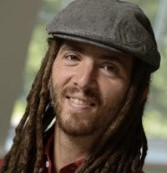}}]{Joshua T. Vogelstein} received the BS degree from the Department of Biomedical Engineering (BME), Washington University, St. Louis, MO, in 2002, the MS degree from the Department of Applied Mathematics \& Statistics (AMS), Johns Hopkins University (JHU) in Baltimore, MD, in 2009, and the PhD degree from the Department of Neuroscience at JHU in 2009. He was a postdoctoral fellow in AMS@JHU from 2009 until 2011, at which time he was appointed an assistant research scientist, and became a member of the Institute for Data Intensive Science and Engineering. He spent two years at Information Initiative at Duke, before coming home to his current appointment as assistant professor in BME@JHU, and core faculty in both the Institute for Computational Medicine and the Center for Imaging Science. He also holds joint appointment in the AMS, Neuroscience, and Computer Science departments at JHU. His research interests primarily include computational statistics, focusing on big data, wide data, and icky data, especially connectomics. His research has been featured in a number of prominent scientific and engineering journals and conferences including Annals of Applied Statistics, IEEE PAMI, NIPS, SIAM Journal of Matrix Analysis and Applications, Science Translational Medicine, Nature Methods, and Science
\end{IEEEbiography}

\begin{IEEEbiography}[{\includegraphics[width=1in,height=2in,clip,keepaspectratio]{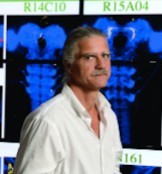}}]{Carey E. Priebe}
received the BS degree in mathematics from Purdue University in 1984, the MS degree in computer science from San Diego State University in 1988, and the PhD degree in information technology (computational statistics) from George Mason University in 1993. From 1985 to 1994 he worked as a mathematician and scientist in the US Navy research and development laboratory system. Since 1994 he has been a professor in the Department of Applied Mathematics and Statistics, Whiting School of Engineering, Johns Hopkins University, Baltimore, Maryland. At Johns Hopkins, he holds joint appointments in the Department of Computer Science, the Department of Electrical and Computer Engineering, the Center for Imaging Science, the Human Language Technology Center of Excellence, and the Whitaker Biomedical Engineering Institute. He is a past president of the Interface Foundation of North America-Computing Science and Statistics, a past chair of the American Statistical Association Section on Statistical Computing, a past vice president of the International Association for Statistical Computing, and is on the editorial boards of the Journal of Computational and Graphical Statistics, Computational Statistics and Data Analysis, and Computational Statistics. His research interests include computational statistics, kernel and mixture estimates, statistical pattern recognition, statistical image analysis, dimensionality reduction, model selection, and statistical inference for high-dimensional and graph data. He is a senior member of the IEEE, a Lifetime Member of the Institute of Mathematical Statistics, an Elected Member of the International Statistical Institute, and a fellow of the American Statistical Association.
\end{IEEEbiography}

\vfill

%\newpage


%\vfill


\end{document}